\documentclass{article}
\usepackage{ijcai22}

\usepackage{times}
\usepackage{soul}
\usepackage{url}
\usepackage[hidelinks]{hyperref}
\usepackage[utf8]{inputenc}
\usepackage[small]{caption}
\usepackage{graphicx}
\usepackage{amsmath}
\usepackage{amsthm}
\usepackage{booktabs}
\usepackage{algorithm}
\usepackage{algorithmic}
\usepackage{bigstrut}
\usepackage{multirow}
\usepackage{multirow}
\usepackage{bigstrut}
\usepackage{graphicx}
\usepackage{longtable}
\usepackage{subfigure}
\usepackage{diagbox}
\title{FRIB: Low-poisoning Rate Invisible Backdoor Attack based on Feature Repair}

\author{
Hui Xia\and
Xiugui Yang\and
Xiangyun Qian\and
Rui Zhang
\affiliations
College of Computer Science and Technology, Ocean University of China, Qingdao 266100, China\\
\emails
\{zhang\_rui0504)\}@163.com
}

\begin{document}

\maketitle

\begin{abstract}
During the generation of invisible backdoor attack poisoned data, the feature space transformation operation tends to cause the loss of some poisoned features and weakens the mapping relationship between source images with triggers and target labels, resulting in the need for a higher poisoning rate to achieve the corresponding backdoor attack success rate. To solve the above problems, we propose the idea of feature repair for the first time and introduce the blind watermark technique to repair the poisoned features lost during the generation of poisoned data. Under the premise of ensuring consistent labeling, we propose a low-poisoning rate invisible backdoor attack based on feature repair, named FRIB. 
Benefiting from the above design concept, the new method enhances the mapping relationship between the source images with triggers and the target labels, and increases the degree of misleading DNNs, thus achieving a high backdoor attack success rate with a very low poisoning rate. Ultimately, the detailed experimental results show that the goal of achieving a high success rate of backdoor attacks with a very low poisoning rate is achieved on all MNIST, CIFAR10, GTSRB, and ImageNet datasets. 

\end{abstract}

\section{Introduction}
A lot of recent research has shown that deep neural networks are vulnerable to backdoor attack. Backdoor attack \cite{li2020backdoor} is a class of attacks that spoof a model by modifying its parameters or using preprocessed inputs. The current mainstream backdoor attack methods are poisoning-based backdoor attacks, which occur in the data collection phase and are implemented by poisoning the training set. It is further subdivided into two main categories: trigger-visible backdoor attacks and invisible backdoor attacks. Trigger-visible backdoor attacks \cite{gu2017badnets} \cite{liu2017trojaning} generate poisoned data by attaching the trigger directly to the source image, which is marked and difficult to apply in practice. Backdoor triggers with invisible features have received significant academic attention in recent years. The invisible backdoor attack with inconsistent labels \cite{chen2017targeted} \cite{turner2019label} \cite{zhong2020backdoor} \cite{bagdasaryan2021blind} \cite{li2020invisible} is that the triggers in the generated poisoned data mostly exist in the form of scrambling or noise, and although the triggers are invisible, such attacks are difficult to evade manual visual inspection. To evade manual visual inspection, \cite{saha2020hidden} proposed label-consistent invisible backdoor attacks, which generate label-consistent poisoned data mainly through feature space transformation. Such methods are less efficient in generating poisoned data and require a high poisoning rate to achieve a high success rate of backdoor attacks. We investigate this kind of poisoning-based backdoor attack in-depth and try to explore the reason why label-consistent invisible backdoor attacks require a high poisoning rate and explore a novel invisible backdoor attack scheme with a low poisoning rate.

	Extensive analysis of the Hidden \cite{saha2020hidden} method led us to an important finding that the poisoned data generated by the feature space transformation lost some of the poisoned features, which led to the DNN's inability to learn these important features, thus weakening the mapping relationship between the source image with the trigger and the target label, and weakening the misleading ability of the DNN, which eventually led to its need for a higher poisoning rate to achieve the corresponding success rate of the backdoor attack. Based on this finding, we ponder whether the above problem can be solved using feature repair? One potential approach is to use image watermarking techniques to repair this lost poisoned feature. We then tried two common image watermarking techniques, plaintext watermarking and blind watermarking. Among them, the blind watermarking technique is more stealthy and can remain invisible to the human eye, so we prefer it. In addition, we also found that the initial pairing of source and target images is required in the feature space transformation process, resulting in low efficiency in generating poisoned data. To address this issue, we wish to bypass the process of the initial pairing of source and target images by adding constraints to the iterative process.

\begin{figure*}[t]
\centering            	
\includegraphics[width=14cm]{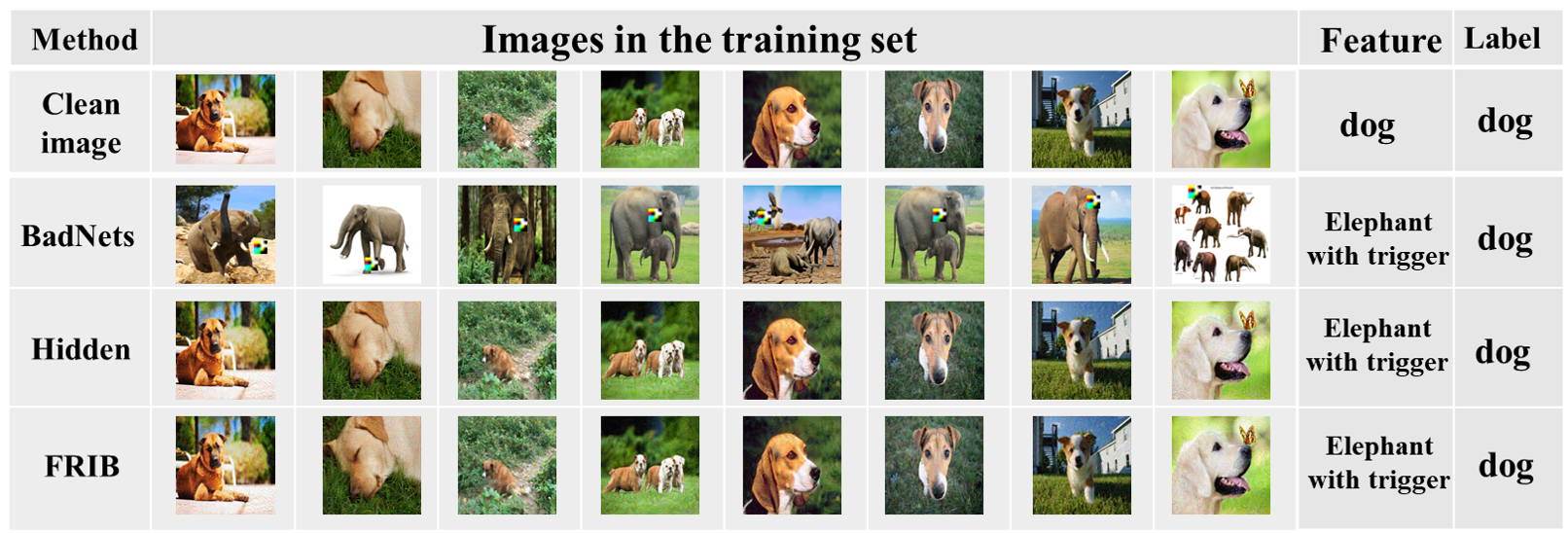}        	
\caption{ Comparison of the poisoned data generated by FRIB with BadNets and Hidden. In the poisoned data generated by BadNets, the feature space is the elephant with the trigger, the label is the dog, and the human eye vision is the elephant with the trigger. In the poisoned data generated by FRIB and Hidden, the feature space is the elephant with the trigger, the label is the dog, and the human eye vision is the dog.}
\end{figure*}

Based on the above assumptions, we add constraints to the feature space transformation algorithm and use the blind watermark technique based on discrete wavelet transformation to repair the features of the poisoned data after the feature space transformation under the premise of ensuring the consistent labeling of the poisoned data. For the first time, we present a low-poisoning rate invisible backdoor attack based on feature repair(FRIB). Figure 1 shows the comparison of FRIB with BadNets and Hidden generated poisoned data.

	We have performed a comprehensive evaluation of the proposed scheme, which achieves the goal of achieving a high success rate of backdoor attacks with a very low poisoning rate on MNIST, CIFAR10, GTSRB, and ImageNet datasets. The main contributions are as follows:
\begin{itemize}
\item  In the invisible backdoor attack, we propose the idea of feature repair for the first time, and creatively propose a low poisoning rate invisible backdoor attack based on feature repair(FRIB). To the best of our knowledge, this is the first label-consistent invisible backdoor attack scheme that uses the blind watermark technique based on discrete wavelet transform for feature repair of poisoned data.

\item  Since the scheme performs feature repair on the poisoned data after feature space transformation and enhances the features of the source images with triggers, the final generated poisoned data has a strong mapping relationship with the target labels, thus achieving a high success rate of backdoor attacks with a very low poisoning rate. In addition, by adding constraints to the iterations, we optimize the feature space transformation algorithm. Since the initial pairing of source and target images is no longer required, the generation efficiency of poisoning data is improved.

\item To verify the effectiveness of the above scheme, we compared FRIB with Hidden and BadNets on MNIST, CIFAR10, GTSRB, and ImageNet datasets. The experimental results show that FRIB not only has a higher attack success rate with the same poisoning rate compared to Hidden but also has a 26.0\% improvement in the efficiency of generating valid poisoned data. Compared with BadNets, FRIB has comparable or even better attack effectiveness under the same poisoning rate. For example, FRIB only needs a 1.16\% poisoning rate on the MNIST dataset to achieve a 100\% attack success rate, and the impact of the model on the original task is negligible.
\end{itemize}

\section{Related Work}
This section describes the current state of research on backdoor attacks in terms of two aspects of poisoning-based backdoor attacks: trigger-visible backdoor attacks and invisible backdoor attacks.

\subsection{Trigger-visible backdoor attacks}
BadNets stands for trigger-visible backdoor attacks, and almost all poisoning-based backdoor attacks are performed based on BadNets. \cite{gu2017badnets} first introduced backdoor attacks into deep learning and proposed a BadNets method to inject backdoors by poisoning some training samples. BadNets labeled some training images with trigger patterns and changed their labels to the target labels and then trained DNNs using the modified poisoned training set. \cite{liu2017trojaning} proposed a Trojan attack against neural networks that uses model inversion to obtain input samples without accessing the training set. It generated triggers by maximizing the activation of certain internal neurons in the model and then retrained the model using reverse-engineered training data. Model retraining created stronger connections between the triggers and a small set of internal neurons, injecting malicious behavior into the model and eventually leading to misclassification. However, because the triggers are so obvious, this trigger-visible backdoor attack is easily detected, making it infeasible in practice.

\subsection{Invisible backdoor attacks}
\cite{chen2017targeted} first discussed the stealthiness of poisoning-based backdoor attacks in terms of the visibility of backdoor triggers. To evade inspection, poisoned data and benign data should be difficult to distinguish. Therefore, they proposed a hybrid strategy to generate poisoned data by mixing backdoor triggers with benign data. They demonstrated that employing a small amplitude random noise as a backdoor trigger can also successfully create backdoors. \cite{turner2019label} proposed to use the backdoor trigger amplitude to perturb the pixel values of benign data instead of replacing the corresponding pixels. \cite{zhong2020backdoor} used a universal adversarial attack \cite{moosavi2017universal} to generate a backdoor trigger that minimizes the L2 parametrization of the perturbation. \cite{bagdasaryan2021blind} considered a backdoor attack as a special multitask optimization to achieve stealth by poisoning loss calculation. \cite{li2020invisible} introduced steganography to backdoor attacks by steganographically embedding the trigger into DNN and thus achieving stealth. \cite{saha2020hidden} proposed a backdoor attack with hidden triggers, where poisoned data is correctly labeled and the trigger is not can be seen. Although such schemes make the triggers invisible, they are inefficient in generating poisoned data and have a low success rate of the attack, requiring a large number of poisoned data to train the target model for the neural network to identify the error.

Inspired by the above scheme, we try to propose a new label-consistent invisible backdoor attack scheme, i.e., under the condition that the poisoned data are correctly labeled, the mapping relationship between source and target images with triggers is enhanced by feature repair to generate poisoned data simply and efficiently, making it possible to poison the whole network with very little poisoned data.
\section{Threat Model Definition}

The classifier is denoted as ${f_w}:X \to {[0,1]^{|Y|}}$ \cite{li2020backdoor}, $X \subset {R^d}$ is the instance space, and $Y = \left\{ {1,2,...,M} \right\}$ is the label space. $f(x)$ denotes the posterior vector with respect to class $M$, $C(x) = \arg \max {f_w}(x)$ denotes the predicted label, ${y_t}$ denotes the target label.

The standard risk $R_{s}$: measures whether the predicted label is the same as the true label, i.e., the model test accuracy.
\begin{equation}R{}_s({D_L}) = {E_{(x,y) \sim {P_{DL}}}}\left[ {{\rm I}\left\{ {C(x) \ne y} \right\}} \right] .\end{equation}
The backdoor risk $R_{b}$: measures whether the trigger can successfully activate the backdoor in the classifier, i.e., the backdoor attack success rate.
\begin{equation}R{}_b({D_L}) = {E_{(x,y) \sim {P_{DL}}}}\left[ {{\rm I}\left\{ {C(x{'}) \ne {y_t}} \right\}} \right] .\end{equation}
The perceivable risk ${R_{p}}$: measures whether the poisoned data can be detected by humans or machines, i.e., whether the label of poisoned data is consistent with its true label. \begin{equation}R{}_p({D_L}) = {E_{(x,y) \sim {P_{DL}}}}\left[ {D(x')} \right] .\end{equation}

Objective functions: \begin{equation}{\min _w} Rs({D_L} - {D_{sL}}) + {\lambda _1} \cdot {R_b}({D_{sL}}) + {\lambda _2} \cdot {R_p}({D_{sL}}),\end{equation}

where $w$ is a model parameter, ${\lambda _1}$, ${\lambda _2}$ is two non-negative trade-off hyperparameters, ${D_L} = \left\{ {({x_i},{y_i})|i = 1,...,{N_l}} \right\}$ denotes the labeled data set, and ${D_{sL}} = \left\{ {(x,y) \subset {D_L}} \right\}$ denotes a subset of ${D_L}$, and $\frac{{|{D_{sL}}|}}{{|{D_L}|}}$ is the poisoning rate. $I( \cdot )$ is the Boolean function, if `$\cdot$' is true, the result of $I( \cdot )$ is 1.
\begin{algorithm}[tb]
\setlength{\tabcolsep}{5mm}{
\caption{Initial poisoned images generation Algorithm}
\label{alg:algorithm}
\textbf{Input}: $s,p,s',t$\\
\textbf{Parameter}: $\lambda,\mu ,epochs$\\
\textbf{Output}: $z$
\begin{algorithmic}[1]
\STATE $s' = s \odot p$
\STATE $x = t$
\STATE ${L_2}\left( x \right) = \parallel f(x) - f(s')\parallel _2^2$
\WHILE{$loss$ is large and $i \le epochs$}
\STATE ${x_i}' = {x_{i - 1}} - \lambda {\Delta _x}{L_2}\left( {{x_{i - 1}}} \right)$
\STATE $||x - {x_{i - 1}}||_2^2{\rm{ }} \le {\rm{ }}{\mu _i}$
\ENDWHILE
\STATE $z = \arg {\min _x}\parallel f(x) - f(s')\parallel _2^2 + \mu \parallel x - t\parallel _2^2$
\STATE return $z$
\end{algorithmic}
}
\end{algorithm}
\section{FRIB}
Our scheme is divided into three main parts, which are: feature space transformation, add blind watermark, and model retraining.
\subsection{Feature Space Transformation}
Optimize the initial poisoned data $z$ by solving the following objective function:
\begin{equation}
\begin{array}{l}
 z = \arg \mathop {\min }\limits_x \parallel f(x) - f(s')\parallel _2^2 + \mu \parallel x - t\parallel _2^2 , \\
 s.t.,||f(x) - f(s')|{|_2} \le \sigma {\rm{   and  }}||x - t|{|_2}{\rm{ }} \le {\rm{ }}\mu , \\
 \end{array}
\end{equation}
where $s$ is the source image, $s'$ is the source image with the trigger $p$, $t$ is the target image and $z$ is the initial poisoned data, ${\left\| {{\rm{ }} \cdot {\rm{ }}} \right\|_2}$ is the ${{\rm{L}}_2}$ norm, $\mu $ is a parameter for balancing, adjusting the coefficient $\mu $ makes the initial poisoned data visually closer to the target image and less likely to be detected.
Let $f\left( x \right)$ denote the function that propagates $x$ through the neural network to the penultimate layer, and call the activation function of this layer the feature space representation of the input. The left side of the objective function is to make the feature space of the initial poisoned data similar to the source image with triggers, and the right side is to make the initial poisoned data visually similar to the target image.

In the step $i$ of the optimization algorithm, we want to obtain the initial poisoned data with a consistent label and follow the following constraint:
\begin{itemize}
\item The initial poisoned data is located in the input domain: ${x_i} + {\eta _i} \in [0,255]$.

\item Relative size per change in the feature space: $||f(x) - f({x_{i - 1}})|{|_2} \le {\sigma _i}$.

\item Changing this size in the feature space makes the relative size visually change: $||x - {x_{i - 1}}|{|_2}{\rm{ }} \le {\rm{ }}{\mu _i}$.
\end{itemize}

Algorithm 1 is the initial poisoned data generation algorithm in which we use the forward-backward-splitting iterative algorithm \cite{goldstein2014field}. First, we optimize $||f(x) - f({x_{i - 1}})|{|_2}$, using gradient descent, and then adjust the coefficients $\mu $ so that they satisfy the constraints in $||x - {x_{i - 1}}|{|_2}{\rm{ }} \le {\rm{ }}{\mu _i}$, and then iterate.

\subsection{Add Blind Watermark}

Under the premise of consistent labeling, in order to reduce the poisoning rate and improve the success rate of backdoor attacks, we use the blind watermark technique based on discrete wavelet transform \cite{al2007combined}\cite{anand2020improved} to repair the features of the poisoned data after feature space transformation, enhancing the features of the source images with triggers.

\subsubsection{Discrete Wavelet Transforms}

The function $\varphi \left( t \right)$ that satisfies the admissibility condition ${{\rm{C}}_\varphi } = {\int_{ - \infty }^\infty  {{{\left| {\varphi \left( \omega  \right)} \right|}^2}\left| \omega  \right|} ^{ - 1}}d\omega  <  + \infty $ is called the proto-wavelet or mother wavelet function $\varphi \left( t \right)$. A family of functions can be generated by translating or stretching the mother wavelet function:
\begin{equation}
\left\{ {{\varphi _{a,b}}\left( t \right)} \right\}:{\varphi _{a,b}}\left( t \right) = \frac{1}{{\sqrt {\left| a \right|} }}\varphi \left( {\frac{{t - b}}{a}} \right){\rm{  }}a,b \in R{\rm{ }}, {\rm{ }}a \ne 0 ,
\end{equation}
${\varphi _{a,b}}\left( t \right)$ is called a wavelet function dependent on the parameters \emph{a}, \emph{b}, referred to as continuous wavelet or wavelet. The wavelet discretization can form a canonical orthogonal basis in ${L^2}\left( R \right)$ space, with which the signal is represented or approximated, and the exact approximation of many different images can be obtained with a smaller number of wavelet coefficients.

The wavelet transform is the inner product of signal $f\left( t \right)$ and wavelet function ${\varphi _{a,b}}\left( t \right)$ with good localization properties, while orthogonal wavelet transform is the same orthogonal transform kernel for both high-pass and low-pass channels, i.e., a standard orthogonal transform is used for each channel. If we let ${C_{0,n}}\left( f \right)$ be the original signal, the orthogonal wavelet decomposition equation is:
\begin{equation}
\begin{array}{l}
 {c_{m,n}}\left( f \right) = \sum\limits_{k \in Z} {h\left( {2n - k} \right){c_{m - 1,k}}\left( f \right)} , \\
 {d_{m,n}}\left( f \right) = \sum\limits_{k \in Z} {g\left( {2n - k} \right){c_{m - 1,k}}\left( f \right)} , \\
 \end{array}
\end{equation}
where \emph{g} and \emph{h} denote the high-pass filter and low-pass filter correspondingly, which should satisfy the relation: $g\left( l \right) = {\left( { - 1} \right)^l}h\left( { - l + 1} \right)$.

The signal reconstruction is given by:
\begin{equation}
{c_{m - 1,l}}\left( f \right) = \sum\limits_{n \in Z} {\left[ {h\left( {2n - l} \right){c_{m,n}}\left( f \right) + g\left( {2n - l} \right){d_{m,n}}\left( f \right)} \right]} .
\end{equation}

Since most of the orthogonal wavelet bases are infinite branches, the filters \emph{h} and \emph{g} associated with this are infinite impulse responses and computationally infeasible. Therefore, for the dual orthogonal wavelet decomposition, the decomposition process can still be used in Eq. (7), but the reconstruction process becomes:
\begin{equation}
{c_{m - 1,l}}\left( f \right) = \sum\limits_{n \in Z} {\left[ {\overline h \left( {2n - l} \right){c_{m,n}}\left( f \right) + \overline g \left( {2n - l} \right){d_{m,n}}\left( f \right)} \right]} ,
\end{equation}
where \emph{h} and $\overline h $ represent the low-pass analytical filter and the low-pass synthetic filter respectively, g and $\overline g $ represent the high-pass analytical filter and high-pass synthetic filter respectively, and $\overline h $ is orthogonal to \emph{g} and $\overline g $ is orthogonal to \emph{h}, i.e., $\overline g \left( n \right) = {\left( { - 1} \right)^n}h\left( {1 - n} \right)$, $g\left( n \right) = {\left( { - 1} \right)^n}\overline h \left( {1 - n} \right)$, $\sum\limits_{n \in Z} {h\left( n \right)\overline h \left( {n + 2k} \right)}  = {\sigma _{k,0}}$.

Since the image is a two-dimensional signal, the image requires a one-dimensional wavelet transform on the rows and columns respectively. The image is discrete wavelet transformed, i.e., decomposed into four quarter-sized subgraphs: horizontal, vertical, and diagonal subgraphs of medium and high-frequency details and low-frequency approximation subgraphs, each obtained by interval sampling filtering. For subsequent decompositions, the approximation subgraphs are decomposed again into smaller subgraphs at the next level of resolution in the same way. By decomposing in this way, the image is decomposed into multiple subgraphs at different resolution levels and in different directions.

\subsubsection{Watermark Information Embedded}
	The original image ${{\bf{I}}_{N*N}}$ and the size of watermark ${{\bf{W}}_{M*M}}$. The original image is the initial poisoned data generated in the previous step, and the watermark is the source image with the trigger.

(\emph{i}){\textbf{Decomposition of the original carrier image}}

The original carrier image ${{\bf{I}}_{N*N}}$ is decomposed into four subgraphs by subsampling:
\begin{equation}
\begin{array}{l}
 {I_1}\left( {{n_1},{n_2}} \right) = I\left( {2{n_1} - 1,2{n_2} - 1} \right),\\
 {I_2}\left( {{n_1},{n_2}} \right) = I\left( {2{n_1} - 1,2{n_2}} \right),\\
 {I_3}\left( {{n_1},{n_2}} \right) = I\left( {2{n_1},2{n_2} - 1} \right),\\
 {I_4}\left( {{n_1},{n_2}} \right) = I\left( {2{n_1},2{n_2}} \right),\\
 \end{array}
\end{equation}

where ${n_1},{n_2} = 0,1, \cdots ,\frac{N}{2} - 1$.

(\emph{ii}){\textbf{Processing of subgraphs and watermarks}}

A two-level wavelet transform is applied to the subgraph ${I_k}$ and the low-frequency approximation coefficients ${V_k}$ are extracted and scanned into a one-dimensional vector in a zigzag fashion. Convert the watermarked image of M*M into a one-dimensional signal $k = 0,1, \cdots ,{M^2} - 1$ by a zigzag method.

(\emph{iii}){\textbf{Embed watermark information}}

The watermark information is embedded at the appropriate position according to the selected sequence of watermark embedding order. For example: embedding a watermark ${W_k}$ on a pair of coefficients at the same position in the wavelet domain of two different subgraphs. First, determine a sequence of watermark embedding to select a bunch of different subgraphs for embedding the watermark. For example, the sequence (1, 2), (3, 4), (3, 1), (4, 2), (3, 2), (4, 1),$ \cdots $, which identifies the two subgraphs (1, 2, 3 or 4) selected for embedding the watermark. The four consecutive numbers in the sequence must be different to ensure that the watermark is embedded in different pairs of subgraphs.

After selecting the coefficient pairs ${V_i}$, ${V_j}$, make the following operations: $V = \frac{{\left( {{V_i} + {V_j}} \right)}}{2}$, if $\left| {\frac{{\left( {{V_i} - {V_j}} \right)}}{V}} \right| < 4\alpha $, then make: ${\overline V _i} = V\left( {1 + \alpha {{\rm{W}}_k}} \right)$, ${\overline V _j} = V\left( {1 - \alpha {{\rm{W}}_k}} \right)$, otherwise do not modify ${V_i}$, ${V_j}$. Where ${\overline V _i}$ and ${\overline V _j}$ are the wavelet coefficients of the image after embedding the watermark, ${W_k}$ is the watermark embedded at that location, and $\alpha $ is the watermark intensity control factor, and the appropriate $\alpha $ is chosen to balance the transparency of the image and the accuracy of the extracted watermark.

(\emph{iv}){\textbf{Compose images with watermarks}}

The coefficient matrix after embedding the watermark is an inverse wavelet transformed to obtain four subgraphs and synthesized to obtain the embedded watermarked image.

\subsection{Model Retraining}
During model retraining, we only fine-tune the parameters of the feature space layer, and we use the RMSprop \cite{zou2019sufficient} gradient descent algorithm to update the parameters during model training. Gradient descent updates the parameters using local gradient information (which can also contain other available information) to gradually approximate the extreme value point of the objective function. The expression is:

\begin{equation}
\begin{array}{l}
 {v_{t + 1}} = \beta {v_t} + (1 - \beta )g_t^2 ,\\
 {\theta _{t + 1}} = {\theta _t} - \frac{\eta }{{\sqrt {{v_{t + 1}} + \varepsilon } }},\\
 \end{array}
\end{equation}
	let ${v_0} = 0$ and obtain:
\begin{equation}
{v_{t + 1}} = (1 - \beta )\sum\limits_{i = 0}^t {{\beta ^{t - i}}} g_i^2 ,
\end{equation}

	where $\eta $ is the learning rate, ${g_t}$ is the gradient depends entirely on the gradient of the current batch, and ${v_{t + 1}}$ is referred to the exponentially weighted mean.

\section{Experimental  Results}

In this section, we introduce the dataset and experimental parameters used for the experiments in Section 5.1. In Section 5.2, by comparing the backdoor attack success rate with BadNets and Hidden, it is verified that FRIB can achieve a high attack success rate with a very low poisoning rate. 
In Section 5.3, we verify that FRIB can generate poisoned data quickly and efficiently by comparing it with Hidden poisoned data generation time.
\subsection{Experimental Setup}
	All experiments are performed on machines with Intel(R) Xeon(R) W-2133 CPUs and NVIDIA GeForce RTX 3090 GPUs using the pytorch machine learning library. We select four commonly used datasets (i.e., MNIST, CIFAR10, GTSRB, ImageNet) for our backdoor attack study, using the fc7 feature embedding f(.) of AlexNet \cite{krizhevsky2017imagenet}. We run Algorithm 1 to generate the initial poisoned data and finally add the blind watermark to generate the final poisoned data. Using small batch gradient descent for 3000 iterations with a batch size of \emph{K}=100 and an initial learning rate of 0.01, we used AlexNet as the basic network with the remaining layer weight parameters unchanged and fine-tuned the parameters of the fc8 layer, and in the model retraining phase, we trained for 100 cycles and used 30 images for testing. For a successful backdoor attack, we expect a high model test accuracy, benign sample accuracy, and a high backdoor attack success rate.

\subsection{Attack Success Rate}
On the MNIST dataset, we choose the source image label as 1 and the target image label as 0. We generate 30 poisoned data with the following numbers of poisoning each time: 5, 10, 15, 20, 25, 30, and add the above different numbers of poisoned data to the training set in the model retraining period to generate the poisoned training set, where the number of clean data in the poisoned training set is 1700.
The model trained using the clean training set: the average test accuracy is 100\%, the average probability that the source image with a trigger is recognized as the target image is 0\%, and the average recognition success rate of the clean image is 100\%. 

On the CIFAR10 dataset, we choose the source image label as the truck and the target image label as the airplane. We generate 60 poisoned data with the following numbers of poisoning each time: 10, 20, 30, 40, 50, 60, where the number of clean data in the poisoned training set is 1000.
The model trained using the clean training set: the average test accuracy is 94.0\%, the average probability that the source image with a trigger is recognized as the target image is 0.90\%, and the average recognition success rate of the clean image is 95.0\%. 

On the GTSRB dataset, we choose the source image label as the speed limit of 80 and the target image label as the parking. We generate 60 poisoned data with the following numbers of poisoning each time: 20, 25, 30, 40, 50, 60, where the number of clean data in the poisoned training set is 1500.
The model trained using the clean training set: the average test accuracy is 100\%, the average probability that the source image with a trigger is recognized as the target image is 0\%, and the average recognition success rate of the clean image is 100\%. 

On the ImageNet dataset, we choose the source image label as the elephant and the target image label as the dog. We generate 40 poisoned data with the following numbers of poisoning each time: 5, 8, 10, 20, 30, 40, where the number of clean data in the poisoned training set is 114.
The model trained using the clean training set: the average test accuracy is 100\%, the average probability that the source image with a trigger is recognized as the target image is 0\%, and the average recognition success rate of the clean image is 100\%. 

Table 1 shows the model test accuracy and benign sample accuracy after retraining with poisoned training sets under MNIST, CIFAR10, GTSRB, and ImageNet datasets. As we can see, FRIB does not affect the usability of the model and only makes the recognition of images with triggers wrong.

To compare the backdoor attack success rate, we compare our scheme (FRIB) with the trigger-visible backdoor attack method BadNets \cite{gu2017badnets} and the label-consistent invisible backdoor attack method Hidden \cite{saha2020hidden}. The experimental results are shown in Figure 2.

\begin{figure}[]
  \includegraphics[width=0.235\textwidth=0,angle=0]{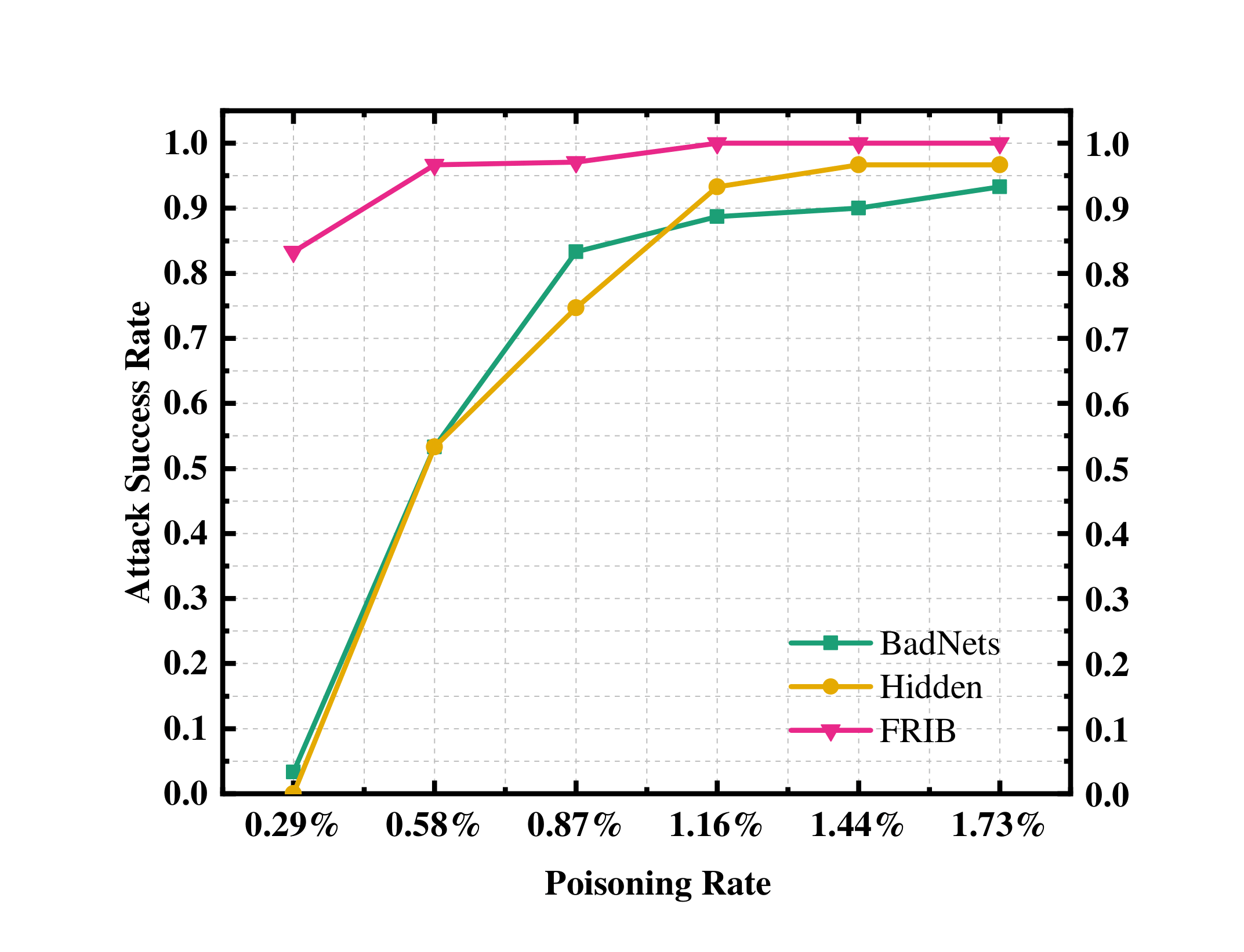}
  \includegraphics[width=0.235\textwidth=0,angle=0]{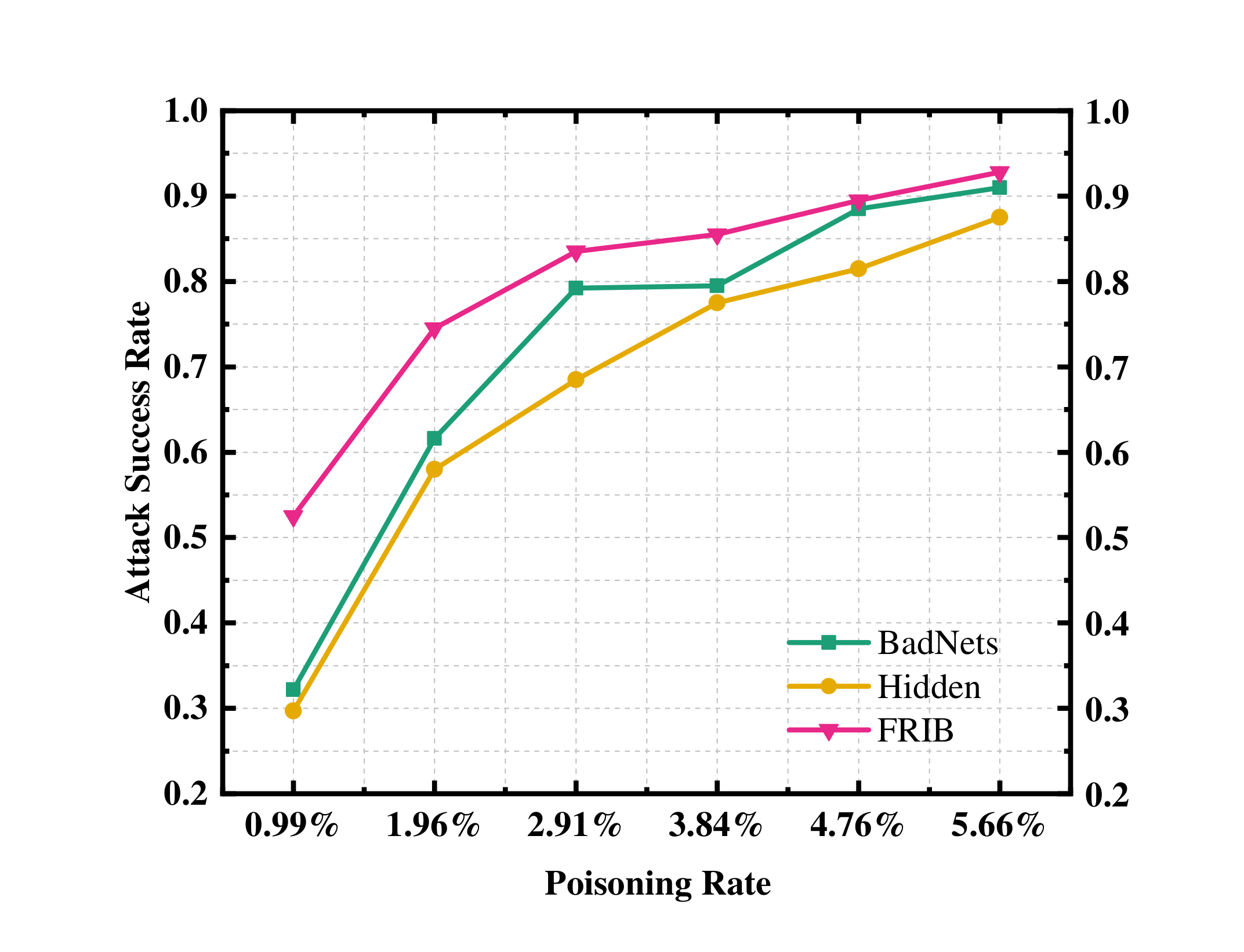}\\
  \scriptsize\indent\qquad\qquad\qquad(\emph{a}) MNIST \qquad\qquad\qquad\qquad\qquad\qquad(\emph{b}) CIFAR10 \\
  \includegraphics[width=0.235\textwidth=0,angle=0]{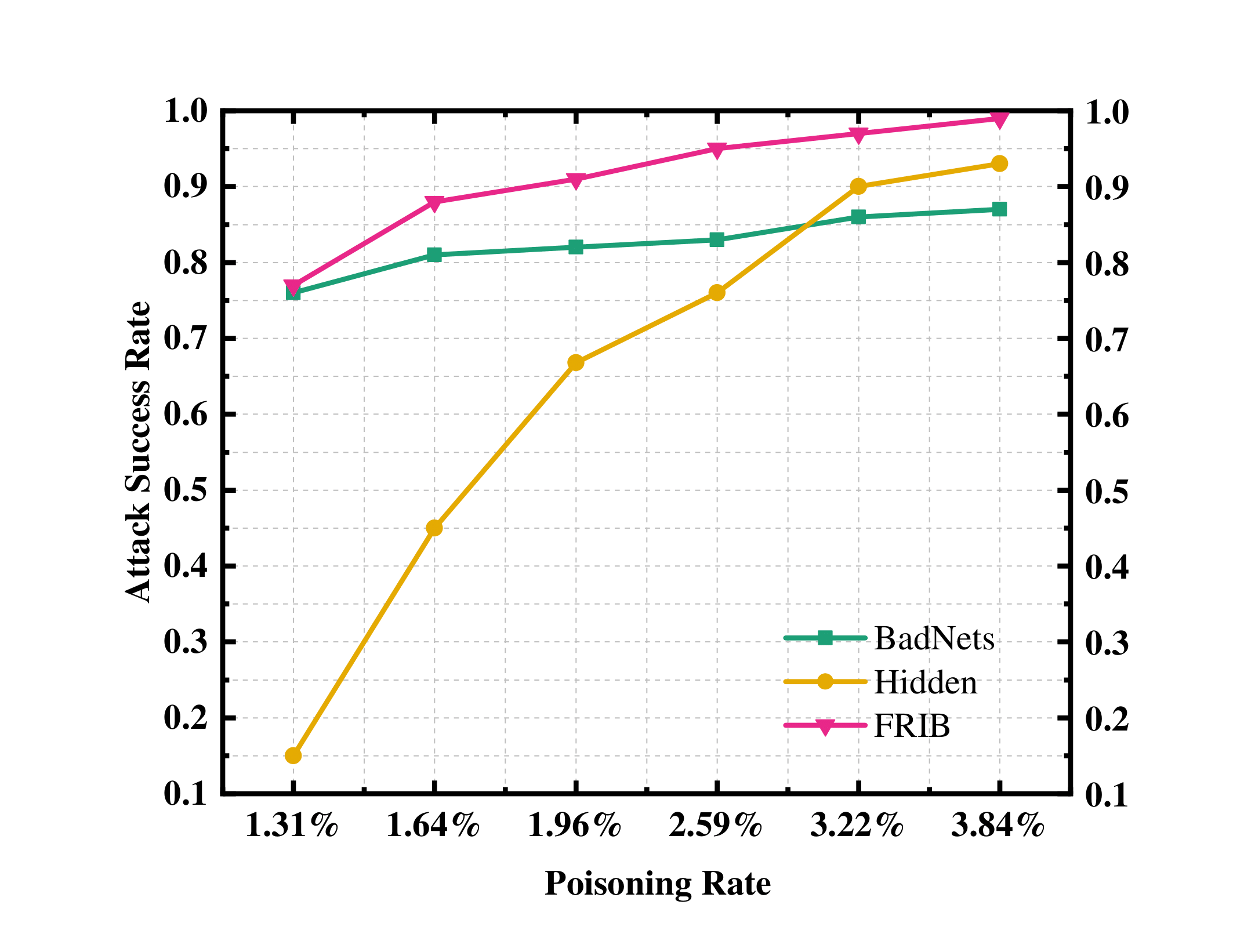}
  \includegraphics[width=0.235\textwidth=0,angle=0]{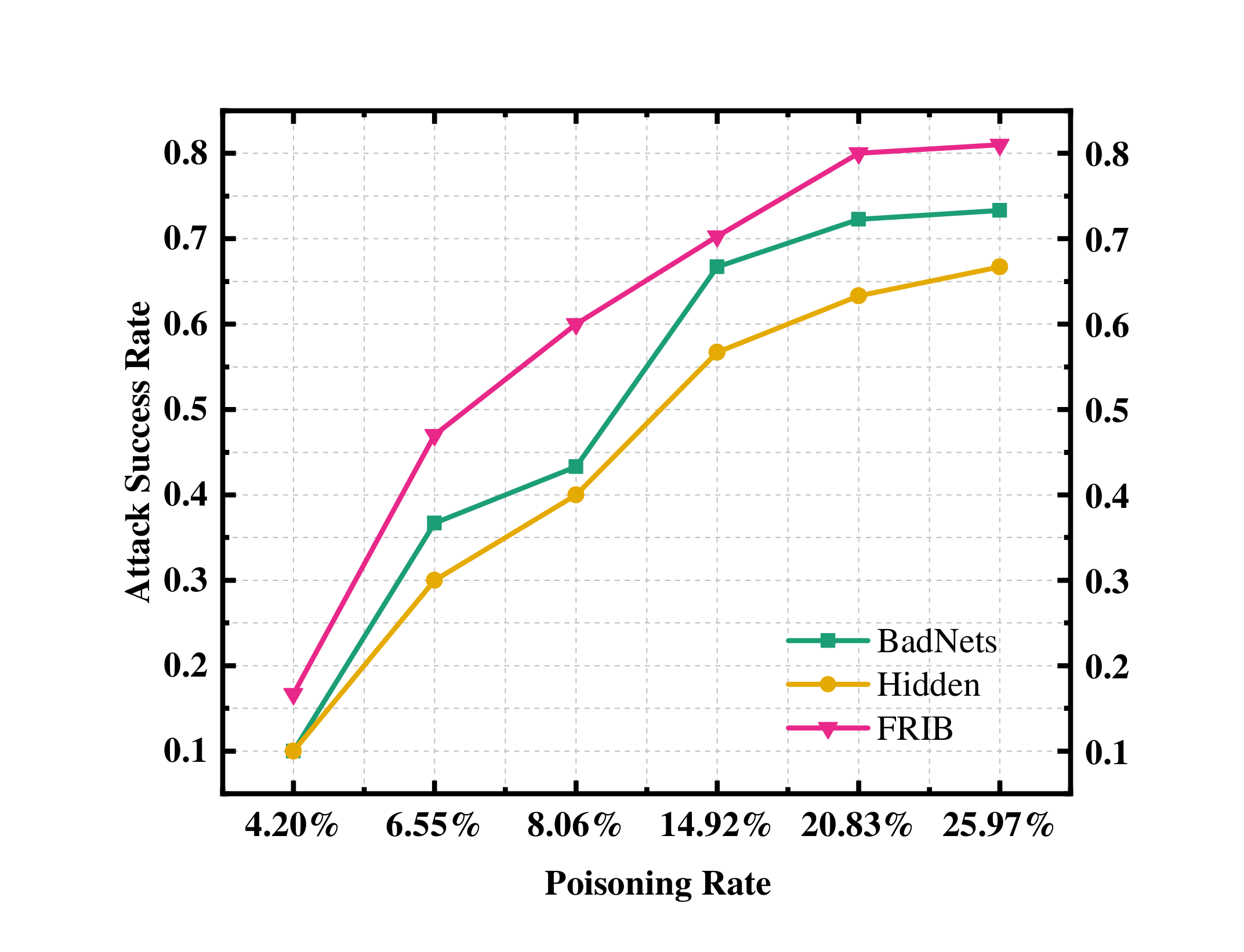}\\
  \scriptsize\indent\qquad\qquad\qquad(\emph{c}) GTSRB \qquad\qquad\qquad\qquad\qquad\qquad(\emph{d}) ImageNet\\
  \caption{Comparison of the backdoor attack success rate.}
\end{figure}

The experimental results show that FRIB backdoor attack success rate is also higher than BadNets in the MNIST dataset with the advantage of invisible triggers. For example, when the proportion of poisoned data is 0.29\%, BadNets backdoor attack success rate is 3.3\%, while FRIB backdoor attack success rate is 83.3\%. FRIB and Hidden scheme are both invisible backdoor attacks with invisible triggers, but it can be seen from the experimental results that the success rate of FRIB backdoor attack is much higher than Hidden under the condition of the same poisoning rate. For example, when the proportion of poisoned data is 0.58\%, the success rate of Hidden backdoor attack is 53.3\%, while the success rate of FRIB backdoor attack is 96.7\%.
In the CIFAR10 dataset, the success rate of FRIB backdoor attack is comparable to BadNets with the advantage of invisible triggers. For example, when the percentage of poisoned data is 5.66\%, the success rate of FRIB backdoor attack is 91.8\%, which is higher than BadNets 91.0\%. For example, when the percentage of poisoned data is 1.96\%, the success rate of Hidden backdoor attack is 58.0\%, while the success rate of FRIB backdoor attack is 74.5\%.
In the GTSRB dataset, with the advantage of invisible triggers, the success rate of FRIB backdoor attack is also higher than BadNets. For example, when the percentage of poisoned data is 1.96\%, the success rate of FRIB backdoor attack is 91.0\%, which is higher than BadNets 82.0\%. When the proportion of poisoned data is 3.84\%, the success rate of FRIB backdoor attack can reach 99.0\%, and the success rate of BadNets backdoor attack is 87.0\% at this time. For example, when the percentage of poisoned data is 1.31\%, the success rate of Hidden backdoor attack is 15.0\%, while the success rate of FRIB backdoor attack is 77.0\%.
In the ImageNet dataset, with the advantage of invisible triggers, the success rate of backdoor attack of FRIB is also higher than that of BadNets. For example, when the percentage of poisoned data is 8.06\%, the success rate of backdoor attack of BadNets is 43.3\%, while the success rate of backdoor attack of FRIB is 60.0\%. For example, when the percentage of poisoned data is 14.92\%, the success rate of Hidden backdoor attack is 56.7\%, while the success rate of FRIB backdoor attack is 70.3\%. 

As shown above, our scheme (FRIB) achieves a high success rate of backdoor attacks with a very low poisoning rate. For example, on the MNIST dataset, FRIB achieves a success rate of 100\% when the percentage of poisoned data is 1.16\%.

\begin{table}[t]
\centering
\small
\renewcommand\tabcolsep{2.0pt}
\caption{Model test accuracy and benign sample accuracy}
\begin{tabular}{lllll}
\toprule
Indicator/Dataset & MNIST & CIFAR10 & GTSRB & ImageNet \\ \midrule
Model   test accuracy            & 100\% & 96.7\%  & 100\% & 100\% \\
Benign sample accuracy           & 100\% & 96.5\%  & 100\% & 100\% \\ \bottomrule
\end{tabular}
\label{tab:booktabs}
\end{table}

\subsection{Efficiency Comparison}
	To examine the efficiency of poisoned data generation, we experimentally compare the poisoned data generation time of Hidden \cite{saha2020hidden}, a label-consistent invisible backdoor attack, with the poisoned data generation time of FRIB.

	In the same experimental environment, we generate 50, 100, 150, 200, 250, and 300 effective poisoned data respectively. Table 2 shows the comparison of the efficiency of poisoned data generated by Hidden and FRIB, and the unit is minute. As seen from the experimental results, the more effective FRIB is as the number of generated poisoned data increases. For example, when generating 50 poisoned data, the generation time of FRIB improved by 13.9\% compared to Hidden, when generating 150 poisoned data, the generation time of FRIB improved by 21.8\% compared to Hidden, and when generating 300 poisoned data, the generation time of FRIB improved by 26.0\% compared to Hidden.
\begin{table}[t]
\centering
\small
\renewcommand\tabcolsep{4.0pt}
\caption{Efficiency comparison of generating poisoned data}
\begin{tabular}{@{}lllllll@{}}
\toprule
Method/Sample Size & 50   & 100   & 150   & 200   & 250   & 300   \\ \midrule
\textbf{FRIB}               & \textbf{7.20}  & \textbf{15.95} & \textbf{21.63} & \textbf{30.18} & \textbf{41.62} & \textbf{48.80} \\
Hidden             & 8.36 & 19.68 & 27.66 & 41.15 & 55.35 & 65.93 \\ \bottomrule
\end{tabular}
\label{tab:booktabs}
\end{table}

\section{Conclusion}
Previous label-consistent invisible backdoor attack schemes mainly face the two problems of low efficiency of poisoned data generation and high poisoning rate of required samples. We introduce the digital blind watermark technique based on discrete wavelet transform and propose an low-poisoning rate invisible backdoor attack based on feature repair (FRIB). Experiments show that our scheme can generate poisoned data quickly and only need very little poisoned data to achieve a high attack success rate.
In addition, we find that the important neurons associated with the target label have a great impact on the backdoor attack success rate. In future research, we will focus on this part of important neurons to explore novel backdoor attack methods and extend them to the real physical world.

\bibliographystyle{plainnat}
\bibliography{reference}

\end{document}